\documentclass{article}
\usepackage[final,nonatbib]{nips_2018}
\usepackage{times}  
\usepackage{helvet}  
\usepackage{courier}  
\usepackage{url}  
\usepackage{graphicx}  

\usepackage{booktabs}       
\usepackage{amsfonts}       
\usepackage{nicefrac}       
\usepackage{microtype}      
\usepackage{lineno}     	

\usepackage{color}
\usepackage{amsmath,amsthm,amssymb,bbold,bm,bbm}
\usepackage{multirow,multicol}

\usepackage{wrapfig}
\usepackage{lipsum}  

\usepackage{algorithm}
\usepackage{algorithmic}

\usepackage{times}
\usepackage{subfigure} 

\usepackage{epstopdf}
\usepackage{enumerate}
\usepackage{makecell}
\usepackage{float}
\usepackage{hhline}

\usepackage{tabularx}
\usepackage{colortbl}

\newcommand{\xv}{{\mathbf x}}
\newcommand{\yv}{{\mathbf y}}
\newcommand{\zv}{{\mathbf z}}
\newcommand{\wv}{{\mathbf w}}

\newcommand{\nv}{{\mathbf n}}

\newcommand{\tv}{{\mathbf t}}
\newcommand{\av}{{\mathbf a}}
\newcommand{\bv}{{\mathbf b}}

\newcommand{\Iv}{{\mathbf I}}

\newcommand{\loss}{{\mathcal L}}
\newcommand{\data}{{\mathcal D}}

\newcommand{\one}{{\mathbf{1}}}
\newcommand{\E}{{\mathbb{E}}}
\newcommand{\Bern}{{\text{Bern}}}
\newcommand{\Norm}{{\mathcal{N}}}

\definecolor{classicrose}{rgb}{0.98, 0.8, 0.91}

\begin{document}
	%
\title{Estimation of Individual Treatment Effect in Latent Confounder Models via Adversarial Learning}
	\author{Changhee Lee,\textsuperscript{1} Nicholas Mastronarde, \textsuperscript{2} and Mihaela van der Schaar \textsuperscript{3, 1} \\
		\textsuperscript{1} Department of Electrical and Computer Engineering, University of California, Los Angeles, USA \\
		\textsuperscript{2} Department of Electrical Engineering, University at Buffalo, New York, USA \\
		\textsuperscript{3} Department of Engineering Science, University of Oxford, UK \\
		chl8856@ucla.edu, nmastron@buffalo.edu, mihaela.vanderschaar@oxford-man.ox.ac.uk}

\maketitle

\begin{abstract}
Estimating the individual treatment effect (ITE) from observational data is essential in medicine.
A central challenge in estimating the ITE is handling confounders, which are factors that affect both an intervention and its outcome.
Most previous work relies on the \textit{unconfoundedness assumption}, which posits that all the confounders are measured in the observational data. 
However, if there are unmeasurable (latent) confounders, then \textit{confounding bias} is introduced.
Fortunately, noisy proxies for the latent confounders are often available and can be used to make an unbiased estimate of the ITE.
In this paper, we develop a novel adversarial learning framework to make unbiased estimates of the ITE using noisy proxies.
\end{abstract}

\section{Introduction}
\begin{figure}[h!]
	\centering
	\begin{minipage}[r]{0.45\textwidth}
		\includegraphics[width=2.0in, trim= 0.1 0.1 0.1 0.1]{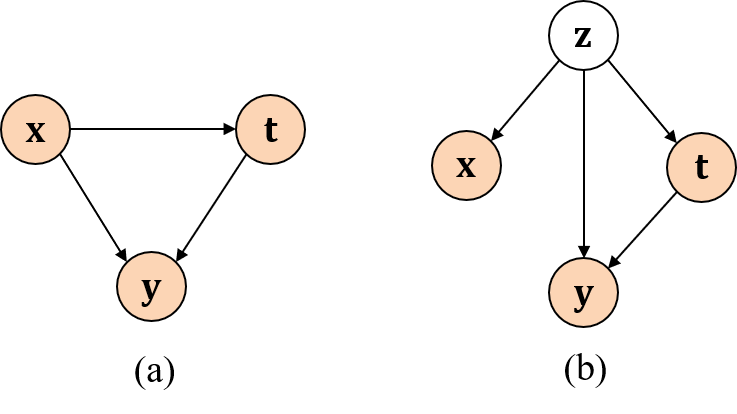}
	\end{minipage}
	\begin{minipage}[l]{0.45\textwidth}
	\caption{Causal diagrams. (a) $\xv$ is an observed confounder between the intervention $\tv$ and outcome $\yv$. (b) $\zv$ is a latent confounder and $\xv$ serves as a proxy that provides noisy views of $\zv$. Shaded and unshaded nodes denote observed and unobserved (latent) variables, respectively.} \label{fig:causal_model}
	\end{minipage}
\end{figure}

Understanding the individual treatment effect (ITE) on an outcome $\yv$ of an intervention $\tv$ on an individual with features $\xv$ is a challenging problem in medicine. When inferring the ITE from observational data, it is common to assume that all of the confounders -- factors that affect both the intervention and the outcome -- are measurable and captured in the observed data as shown in Figure \ref{fig:causal_model}(a). However, in practice, there are often unobserved (latent) confounders $\zv$ as shown in Figure \ref{fig:causal_model}(b). For example, socio-economic status cannot be directly measured, but can influence the types of medications that a subject has access to; therefore, it acts as a confounder between the medication and the patient's health. 
If such latent confounders are not appropriately accounted for, then the estimated ITE will be subject to \textit{confounding bias}, making it impossible to estimate the effect of the intervention on the outcome without bias \cite{Pearl:00,Kuroki:14}.
A common technique for mitigating confounding bias is using \textit{proxy variables}, which are measurable proxy for the latent confounders that can enable unbiased estimation of the ITE. For instance, in the causal diagram in Figure \ref{fig:causal_model}(b), $\xv$ can be viewed as providing noisy proxies of the latent confounders $\zv$.

\textbf{Contributions:} We introduce an adversarial framework to infer the complex non-linear relationship between the proxy variables (i.e., observations) and the latent confounders via approximately recovering posterior distributions that can be used to infer the ITE. 
Our experiments on synthetic and semi-synthetic observational datasets show that the proposed method is competitive with -- and often outperforms -- state-of-the-art methods when there are no latent confounders or when the proxy noise is small, and outperforms all tested benchmarks when the proxy variables become noisy.


\section{Causal Effect with Latent Confounders} \label{sec:causal_effect}
Our goal is to estimate the ITE from an observational dataset $\data = \left\{ (\xv_{i}, t_{i}, \yv_{i}) \right\}_{i=1}^{N}$, where $\xv_{i}$, $t_i$, and $\yv_{i}$, denote the $i$-th subject's feature vector, treatment (we assume that the treatment is binary, i.e., $t \in \{0,1\}$), and outcome vector, respectively, and $N$ is the number of subjects.
The ITE for a subject with observed potential confounder $\xv$ is defined as
\begin{equation} \label{eq:ITE}
ITE(\xv) = \E \left[\yv|\xv, do(t=1) \right] - \E \left[\yv|\xv, do(t=0) \right].
\end{equation}
To recover the ITE under the latent confounder model in Figure \ref{fig:causal_model}(b), we need to identify $p(\yv|\xv, do(t=1))$ and $p(\yv|\xv, do(t=0))$. The former can be calculated as follows:
\begin{equation} \label{eq:probability_outcome}
p(\yv|\xv,\!do(t\!=\!1)) = \! \int_{\zv}\! p(\yv|\zv,\xv,\!do(t\!=\!1)) p(\zv|\xv,\!do(t\!=\!1)) d\zv  = \! \int_{\zv}\! p(\yv|\zv,\xv,t=1) p(\zv|\xv) d\zv, 
\end{equation}
where the second equality follows from the rules of \textit{do}-calculus\footnote{The \textit{do}-operator \cite{Pearl:00} simulates physical interventions by deleting certain functions from the model, replacing them with a constant value, while keeping the rest of the model unchanged.} applied to the causal graph in Figure \ref{fig:causal_model}(b) \cite{Pearl:10}. ($p(\yv|\xv, do(t=0))$ can be derived similarly.) It is worth to highlight that $p(\yv|\xv, do(t=1))$ is equivalent to $p(\yv|\xv, t=1)$ if the unconfoundedness assumption holds as in Figure \ref{fig:causal_model}(a).

Thus, from \eqref{eq:ITE} and~\eqref{eq:probability_outcome}, we can make estimates of the ITE without confounding bias using the estimates of the conditional distributions $p(\yv|\zv, \xv, t)$ and $p(\zv|\xv)$. Since $\zv$ is unobservable, we assume that the joint distribution $p(\zv, \xv, t, \yv)$ can be approximately recovered solely from the observations $(\xv, t, \yv)$ as justified in~\cite{Louizos:17}.

\section{Adversarial Learning for Causal Effect} \label{sec:CEGAN}
In this section, we propose a method that estimates \textit{Causal Effect using a Generative Adversarial Network (CEGAN)}. CEGAN's objective is to estimate the conditional posteriors in \eqref{eq:probability_outcome} under the causal graph in Figure \ref{fig:causal_model}(b) so that we can estimate the ITE~\eqref{eq:ITE} for new subjects. However, since we cannot measure the \textit{true} latent confounder, we are unable to directly learn the posterior distribution $p(\zv|\xv)$. Instead, we learn a mapping between the data (observations) and an arbitrary latent space following an adversarial learning framework similar to those developed in \cite{Dumoulin:17} and \cite{Donahue:17}.
\begin{figure*}[t!]
	\centering 
	\includegraphics[width=0.9\linewidth, trim={0cm 0cm 0cm 0cm}, clip]{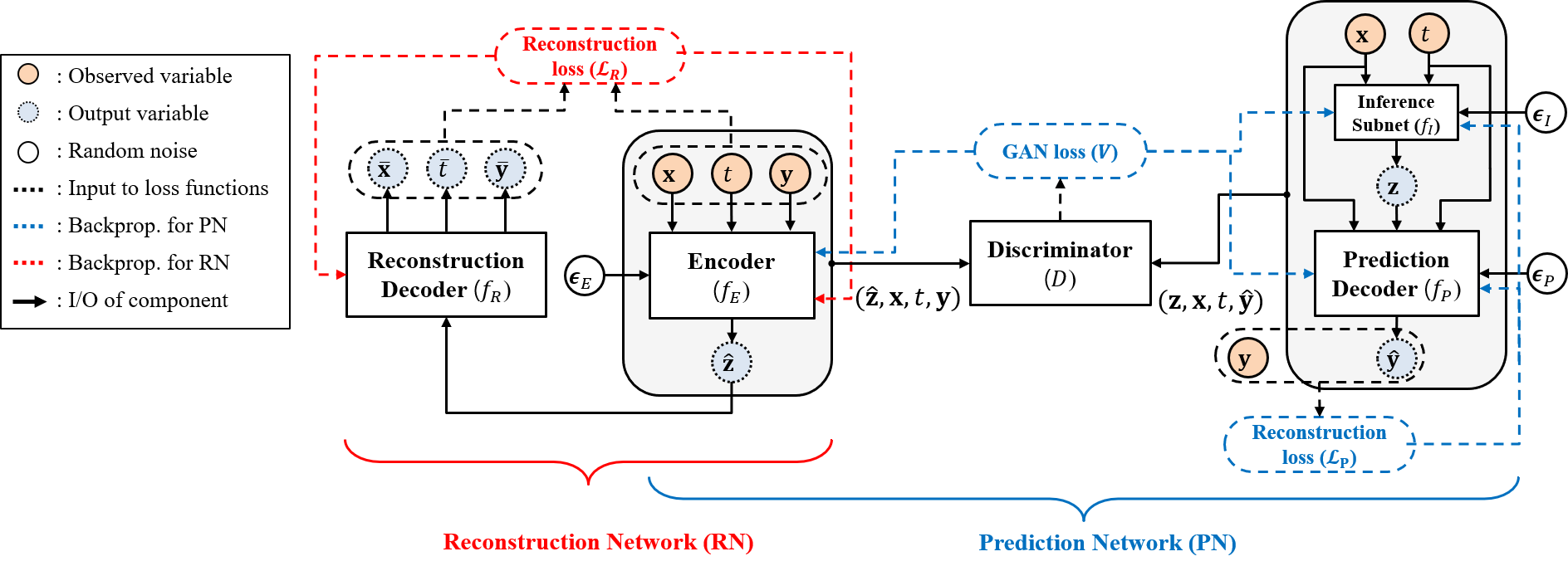}
	\vspace{-1mm}
	\caption{An illustration of the proposed network architecture.}
	\label{fig:network_architecture}
	\vspace{-2.5mm}
\end{figure*}

Our model, depicted in Figure~\ref{fig:network_architecture}, comprises a \textit{prediction network} (right) and a \textit{reconstruction network} (left). Each network includes an encoder-decoder pair, where the encoder is shared between them.
The posterior distributions that are required to solve \eqref{eq:probability_outcome} can be estimated using \textit{bidirectional models} \cite{Dumoulin:17} and \cite{Donahue:17} via factorizing the posterior distribution as 
$p(\yv|\zv,\xv,t) \approx q_{P}(\yv|\zv,\xv,t)$ and $p(\zv|\xv) \approx \sum_{t \in \{0,1\}} q_I(\zv|\xv,t) q(t|\xv)$, where $q_{P}$ and $q_{I}$ are the components of the prediction network and $q$ is the propensity score.
Meanwhile, the reconstruction network is a \textit{denoising autoencoder} \cite{Vincent:10}, which helps the prediction network find a meaningful mapping to the latent space that preserves information in the data space. 

\subsection{Prediction Network}
The prediction network has two components: a \textit{generator} (which consists of the encoder $f_E$, the prediction decoder $f_P$, and the inference subnetwork $f_I$) and a \textit{discriminator} $D$.

The \textbf{encoder} ($f_{E}$), which is employed in both the reconstruction and prediction networks, maps the data space to the latent space. Thus, the output of the encoder $\hat{\zv}$ is given by $\hat{\zv} = f_{E}(\xv, t, \yv, \bm{\epsilon}_{E})$.
The \textbf{inference subnetwork} ($f_I$) is introduced to infer $\zv$ based on $\xv$ and given $t$; its output $\zv$ is given by $\zv = f_{I}(\xv, t, \bm{\epsilon}_{I})$.
The \textbf{prediction decoder} ($f_{P}$) is a function that outputs the estimated outcome $\hat{\yv}$ given a sample $(\xv, t)$ drawn from the data distribution $p_{d}(\xv, t)$ and a latent variable $\zv \sim q_{I}(\zv|\xv, t)$ inferred by $f_I$; thus, $\hat{\yv} = f_{P}(\zv, \xv, t, \bm{\epsilon}_{P})$.
Note that the outputs of the generator are randomized by the noise term $\bm{\epsilon}_{E}, \bm{\epsilon}_{I}, \bm{\epsilon}_{P} \sim \Norm(\mathbf{0}, \Iv)$ using the universal approximator technique described in \cite{Makhzani:16}.

With the conditional probabilities $q_{E}(\zv | \xv, t, \yv)$, $q_{I}(\zv|\xv, t)$, and $q_{P}(\yv|\zv, \xv, t)$ obtained from the generator, we are able to define two joint distributions: $q_{E}(\zv, \xv, t, \yv) = p_{d}(\xv, t, \yv) q_{E}(\zv | \xv, t, \yv)$ for the encoder and $q_{P}(\zv, \xv, t, \yv) = p_{d}(\xv, t)q_{I}(\zv|\xv, t)q_{P}(\yv|\zv,\xv, t)$ for the prediction decoder. 
Using tuples drawn from the two joint distributions, CEGAN attempts to match these distribution by playing an adversarial game between the generator and the discriminator. 
To do so, the \textbf{prediction discriminator} ($D$) maps tuples $(\zv,\xv,t,\yv)$ to a probability in $[0,1]$. Specifically, $D(\zv, \xv, t, \yv)$ and $1 - D(\zv, \xv, t, \yv)$ denote estimates of the probabilities that the tuple $(\zv,\xv,t,\yv)$ is drawn from $q_{E}(\zv, \xv, t, \yv)$ and $q_{P}(\zv, \xv, t, \yv)$, respectively. 
The discriminator tries to distinguish between tuples $(\zv,\xv,t,\yv)$ that are drawn from $q_{E}(\zv, \xv, t, \yv)$ and $q_{P}(\zv, \xv, t, \yv)$. Following the framework in \cite{Dumoulin:17}, the two distributions can be matched (i.e., they reach the same saddle point) by solving the following min-max problem between the generator and the discriminator:
\begin{equation} \label{eq:minmax_prob}
\min_{(\theta_E, \theta_I, \theta_P)} \max_{\theta_{D}}  ~~ \E_{q_{E}(\zv, \xv, t, \yv)} \! \Big[ \log\left(D(\hat{\zv}, \xv, t, \yv)\right)\Big]
+ \E_{q_{P}(\zv, \xv, t, \yv)} \! \Big[ \log\left(1 - D(\zv, \xv, t, \hat{\yv})\right)\Big].
\end{equation}

\subsection{Reconstruction Network}
The relationship between the data and the latent space is not specified in the prediction network. Consequently, the network may converge to an undesirable matched joint distribution. For instance, it may learn to match the joint distributions $q_{E}(\zv, \xv, t, \yv)$ and $q_{P}(\zv, \xv, t, \yv)$ while inferring latent variables $\zv$ that provide no information about the data samples $(\xv, t, \yv)$.
We introduce a reconstruction network to nudge the prediction network toward learning a meaningful mapping between the data and latent spaces. We utilize a denoising autoencoder for the reconstruction network, which employs the same encoder as the prediction network, $f_{E}$, and a \textbf{reconstruction decoder} $f_R$.
$f_{R}$ reconstructs the original input of the encoder $f_{E}$ from the output of $f_{E}$; the output can be given as $(\bar{\xv}, \bar{t}, \bar{\yv}) =  f_{R}(\hat{\zv})$.

Then, we define the following reconstruction loss:
\begin{equation} \label{eq:loss_reconstruction}
\loss_{R}(\wv, \bar{\wv}) = \ell(\xv, \bar{\xv}) + \ell(t, \bar{t}~) + \ell(\yv, \bar{\yv}),
\end{equation}
where $\wv = [\xv, t, \yv]$, $\bar{\wv} = [\bar{\xv}, \bar{t}, \bar{\yv}]$, $\ell(\av, \bv) = \|\av-\bv\|^{2}$ for continuous values, and $\ell(\av, \bv) = - \av^T \log \bv - (\mathbf{1}-\av)^T \log (\mathbf{1}-\bv)$ for binary values. Here, $\log$ denotes the element-wise logarithm.
By minimizing \eqref{eq:loss_reconstruction} iteratively with the min-max problem \eqref{eq:minmax_prob}, $f_{E}$ is able to map data samples into the latent space while preserving information that is available in the data space.

\section{Experiments}\label{sec:experiments}
Ground truth counterfactual outcomes are never available in observational datasets, which makes it difficult to evaluate causal inference methods. 
Thus, we evaluate CEGAN against various benchmarks using a semi-synthetic dataset where we model the proxy mechanism to generate latent confounding. In the appendix, we perform further comparisons using a semi-synthetic dataset suggested in \cite{Louizos:17} and a synthetic dataset.

\textbf{Performance Metric:} We use two different performance metrics in our evaluations -- expected precision in the estimation of heterogeneous effect (PEHE) and average treatment effect (ATE) \cite{Hill:11}:
\begin{equation}\nonumber
\epsilon_{\text{PEHE}} = \frac{1}{N}\! \sum_{i=1}^{N} \! \Big(\! \big(y_{i}(1) - y_{i}(0) \big) - \big(\hat{y}_{i}(1) - \hat{y}_{i}(0)\big)\!\Big)^2, ~~
\epsilon_{\text{ATE}} = \Big| \frac{1}{N}\! \sum_{i=1}^{N} \! \big(y_{i}(1) - y_{i}(0) \big) - \big(\hat{y}_{i}(1) - \hat{y}_{i}(0)\big) \Big|,
\end{equation}
where $y_{i}(1)$ and $y_{i}(0)$ are the ground truth of the treated and controlled outcomes for the $i$-th sample and $\hat{y}_{i}(1)$ and $\hat{y}_{i}(0)$ are their estimates.

\begin{table*}[t!]
	\caption{Comparison of $\sqrt{\epsilon_{\text{PEHE}}}$ and $\epsilon_{\text{ATE}}$ (mean $\pm$ std) on the TWINS dataset.} \label{Table:TWINS_Additional}
	\begin{center}
		\scriptsize
		\setlength{\tabcolsep}{2.5pt}
		\begin{tabular}{|c|c|c|c|c|c|c|c|c|}
			\hline
			\multirow{3}{*}{\textbf{Method}}&\multicolumn{4}{c|}{$\sqrt{\epsilon_{\text{PEHE}}}$} &\multicolumn{4}{c|}{$\epsilon_{\text{ATE}}$} \\ \cline{2-9} 
			&\multicolumn{2}{c|}{\textit{no latent confounding}} &\multicolumn{2}{c|}{\textit{latent confounding}}& \multicolumn{2}{c|}{\textit{no latent confounding}} &\multicolumn{2}{c|}{\textit{latent confounding}} \\\cline{2-9} 
			&In-sample	   &Out-sample   &In-sample    &Out-sample    &In-sample	   &Out-sample   &In-sample    &Out-sample   \\ \hline
			LR-1				&0.365$\pm$0.00  &0.367$\pm$0.00 &0.413$\pm$0.01 &0.423$\pm$0.02&0.045$\pm$0.02&0.186$\pm$0.03&0.064$\pm$0.02&0.206$\pm$0.03 \\
			LR-2				&0.404$\pm$0.02  &0.411$\pm$0.02 &0.442$\pm$0.02 &0.454$\pm$0.02&0.128$\pm$0.03&0.206$\pm$0.04&0.148$\pm$0.03&0.227$\pm$0.04 \\
			kNN   		     	&0.486$\pm$0.02  &0.506$\pm$0.02 &0.492$\pm$0.02 &0.515$\pm$0.02&0.254$\pm$0.04&0.264$\pm$0.04&0.271$\pm$0.04&0.285$\pm$0.04 \\
			CForest 	  	    &\textbf{0.356$\pm$0.01}  &0.372$\pm$0.01 &0.417$\pm$0.02 			&0.429$\pm$0.02&0.025$\pm$0.02&0.188$\pm$0.03&0.023$\pm$0.02&0.186$\pm$0.03 \\
			BART   	     		&0.569$\pm$0.06&0.562$\pm$0.06&0.877$\pm$0.08&0.871$\pm$0.08&0.432$\pm$0.08&0.429$\pm$0.08&0.790$\pm$0.09&0.786$\pm$0.09 \\
			CMGP				&0.367$\pm$0.01&0.365$\pm$0.01&0.430$\pm$0.05&0.438$\pm$0.05&0.034$\pm$0.03&0.036$\pm$0.04&0.192$\pm$0.09&0.213$\pm$0.09 \\
			CFR$_{\text{WASS}}$ &0.371$\pm$0.03&0.371$\pm$0.03&0.427$\pm$0.05&0.438$\pm$0.05&0.056$\pm$0.06&0.071$\pm$0.06&0.205$\pm$0.07&0.226$\pm$0.07 \\
			CEVAE   	   		&0.363$\pm$0.00&0.364$\pm$0.00&0.423$\pm$0.00 &0.428$\pm$0.00&0.071$\pm$0.01&0.165$\pm$0.01&0.088$\pm$0.01&0.183$\pm$0.01 \\ \hline
			CEGAN				&0.363$\pm$0.00  &\textbf{0.362$\pm$0.00} &\textbf{0.369$\pm$0.00} &\textbf{0.369$\pm$0.00}&\textbf{0.018$\pm$0.01}&\textbf{0.017$\pm$0.01}&\textbf{0.022$\pm$0.01}&\textbf{0.021$\pm$0.02} \\ \hline
		\end{tabular}
		\vspace{-1mm}
	\end{center}
\end{table*}

We compare CEGAN against benchmarks (see appendix for details of the tested benchmarks) using a semi-synthetic dataset (\textbf{TWINS}) which is similar to that was first proposed in \cite{Louizos:17}. Based on records of twin births in the USA from 1989-1991 \cite{Almond:05}, we artificially create a binary treatment such that $t=1$ ($t=0$) denotes being born the heavier (lighter). 
The binary outcome is the mortality of each of the twins in their first year. (Since we have records for both twins, we treat their outcomes as two potential outcomes, i.e., $\yv(1)$ and $\yv(0)$, with respect to the treatment assignment of being born heavier.)
Due to its high correlation with the outcome \cite{Moser:07,Platt:14}, we select the feature `\texttt{GESTAT}', which is the gestational age in weeks. 
The treatment assignment is based only on this single variable, i.e., $t_{i}|z_{i} \sim \Bern(\sigma(w z_{i}))$, where $w \sim \Norm(10, 0.1^2)$ and $z$ is the min-max normalized value of `\texttt{GESTAT}'. 
The data generation process is not exactly equivalent to that proposed in \cite{Louizos:17} as i) it includes artificial proxies of the latent variable in the observational dataset, which is less realistic and ii) the treatment assignment is not only based on the latent variable but also on the observed variables which is not consistent with the causal model in Figure \ref{fig:causal_model}(b). (In the appendix, we reported details and results for the TWINS dataset with the same data generation process in \cite{Louizos:17}.)

To assess the performance of causal inference methods in the presence of latent confounding, we test them on two datasets: ``\textit{no latent confounding}'' which contains `\texttt{GESTAT}' and relies on the unconfoundedness assumption as depicted in Figure \ref{fig:causal_model}(a) and ``\textit{latent confounding}'' which excludes `\texttt{GESTAT}' from the observational dataset and follows the latent causal graph in Figure \ref{fig:causal_model}(b). 

Throughout the evaluation, we average over 100 Monte Carlo samples from the estimated posteriors derived using each method to compute $\E (\yv|\xv, do(t=1))$ and $\E (\yv|\xv, do(t=0))$ in \eqref{eq:ITE} for CEVAE and CEGAN. The reported values in Table \ref{Table:TWINS_Additional} are averaged over 50 realizations with the same 64/16/20 train/validation/test splits.

The performance of  $\sqrt{\epsilon_{\text{PEHE}}}$ and $\epsilon_{\text{ATE}}$ is reported in Table \ref{Table:TWINS_Additional}, for both within-sample and out-of-sample tests. The ITE estimation accuracy decreases for all of the evaluated methods after removing `\texttt{GESTAT}' from the observational dataset due to information loss and confounding bias due to the latent confounder. CEGAN provides competitive performance compared to the state-of-the-art when there is no latent confounding, while outperforming all benchmarks under latent confounding for both $\sqrt{\epsilon_{\text{PEHE}}}$ and $\epsilon_{\text{ATE}}$. Under the circumstances when there is latent confounding and the treatment assignment is solely based on this latent confounder, CEGAN provides more robust performance than CEVAE.

\section{Conclusion}\label{sec:conclusion}
In this paper, we studied the problem of estimating causal effects in the latent confounder model. In order to obtain unbiased estimates of the ITE, we introduced a novel method, CEGAN, which utilizes an adversarially learned bidirectional model along with a denoising autoencoder. CEGAN achieves competitive performance with numerous state-of-the-art benchmarks when the \textit{unconfoundedness assumption} holds or the proxy noise is small, while outperforming state-of-the-art causal inference methods when latent confounding is present. CEGAN performs especially well when the proxy noise is large and the treatment is determined based solely on the latent confounders.
\newpage


\bibliographystyle{unsrt}

\begin{thebibliography}{10}
	
	\bibitem{Pearl:00}
	Judea Pearl.
	\newblock {\em {Causality: Models, Reasoning, and Inference}}.
	\newblock Cambridge University Press, 2000.
	
	\bibitem{Kuroki:14}
	Manabu Kuroki and Judea Pearl.
	\newblock Measurement bias and effect restoration in causal inference.
	\newblock {\em Biometrika}, 101(2):423–437, March 2014.
	
	\bibitem{Pearl:10}
	Judea Pearl.
	\newblock On measurement bias in causal inference.
	\newblock {\em In Proceedings of the 26th Conference on Uncertainty in
		Artificial Intelligence (AUAI 08)}, page 425–432, 2010.
	
	\bibitem{Louizos:17}
	Christos Louizos, Uri Shalit, Joris Mooij, David Sontag, Richard Zemel, and Max
	Welling.
	\newblock Causal effect inference with deep latent-variable models.
	\newblock {\em In Proceedings of the 30th Conference on Neural Information
		Processing Systems (NIPS 2017)}, 2017.
	
	\bibitem{Dumoulin:17}
	Vincent Dumoulin, Ishmael Belghazi, Ben Poole, Olivier Mastropietro, Alex Lamb,
	Martin Arjovsky, and Aaron Courville.
	\newblock Adversarially learned inference.
	\newblock {\em In Proceedings of the 5th International Conference on Learning
		Representations (ICLR 2017)}, 2017.
	
	\bibitem{Donahue:17}
	Jeff Donahue, Philipp Krähenbühl, and Trevor Darrell.
	\newblock Adversarial feature learning.
	\newblock {\em In Proceedings of the 5th International Conference on Learning
		Representations (ICLR 2017)}, 2017.
	
	\bibitem{Vincent:10}
	Pascal Vincent, Hugo Larochelle, Isabelle Lajoie, Yoshua Bengio, and
	Pierre-Antoine Manzagol.
	\newblock Stacked denoising autoencoders: Learning useful representations in a
	deep network with a local denoising criterion.
	\newblock {\em Journal of Machine Learning Research}, 11:3371--3408, 2010.
	
	\bibitem{Makhzani:16}
	Alireza Makhzani, Jonathon Shlens, Navdeep Jaitly, Ian Goodfellow, and Brendan
	Frey.
	\newblock Adversarial autoencoders.
	\newblock {\em Proc. 2016}, 2016.
	
	\bibitem{Hill:11}
	Jennifer~L. Hill.
	\newblock Bayesian nonparametric modeling for causal inference.
	\newblock {\em Journal of Computational and Graphical Statistics},
	20(1):217–240, 2011.
	
	\bibitem{Almond:05}
	Douglas Almond, Kenneth~Y. Chay, and David~S. Lee.
	\newblock The costs of low birth weight.
	\newblock {\em The Quarterly Journal of Economics}, 120(3):1031–1083, 2005.
	
	\bibitem{Moser:07}
	Kath Moser, Alison Macfarlane, Yuan~Huang Chow, Lisa Hilder, and Nirupa
	Dattani.
	\newblock Introducing new data on gestation-specific infant mortality among
	babies born in 2005 in england and wales.
	\newblock {\em Health Stat Q.}, 35:13--27, 2007.
	
	\bibitem{Platt:14}
	M.~J. Platt.
	\newblock Outcomes in preterm infants.
	\newblock {\em Health Stat Q.}, 128(5):399--403, 2014.
	
	\bibitem{Crump:08}
	Richard~K. Crump, V.~Joseph Hotz, Guido~W. Imbens, and Oscar~A. Mitnik.
	\newblock Nonparametric tests for treatment effect heterogeneity.
	\newblock {\em The Review of Economics and Statistics}, 90(3):389–405, 2008.
	
	\bibitem{Chipman:10}
	Hugh~A. Chipman, Edward~I. George, and Robert~E. McCulloch.
	\newblock Bart: Bayesian additive regression trees.
	\newblock {\em The Annals of Applied Statistics}, 4(1):266--298, 2010.
	
	\bibitem{Wager:17}
	Stefan Wager and Susan Athey.
	\newblock Estimation and inference of heterogeneous treatment effects using
	random forests.
	\newblock {\em Journal of the American Statistical Association}, 2017.
	
	\bibitem{Shalit:16}
	Uri Shalit, Fredrik~D. Johansson, and David Sontag.
	\newblock Estimating individual treatment effect: Generalization bounds and
	algorithms.
	\newblock {\em In Proceedings of the 33rd International Conference on
		International Conference on Machine Learning (ICML 2016)}, 2016.
	
	\bibitem{Ahmed:NIPS17b}
	Ahmed~M Alaa and Mihaela van~der Schaar.
	\newblock Bayesian inference of individualized treatment effects using
	multi-task gaussian processes.
	\newblock {\em In Proceedings of the 30th Conference on Neural Information
		Processing Systems (NIPS 2017)}, 2017.
	
	\bibitem{allman2009identifiability}
	Elizabeth~S Allman, Catherine Matias, and John~A Rhodes.
	\newblock Identifiability of parameters in latent structure models with many
	observed variables.
	\newblock {\em The Annals of Statistics}, pages 3099--3132, 2009.
	
\end{thebibliography}

\clearpage

\makebox[\textwidth]{\Large \textbf{Appendix}}
\vspace{3mm}

\section{Optimization of CEGAN}\label{sec:training}
CEGAN is trained in an iterative fashion: we alternate between optimizing the reconstruction network and the prediction network until convergence. In this section, we describe the empirical loss functions that are used to optimize each component. Pseudo-code for training CEGAN is provided in Algorithm \ref{alg:learning_algo}.

{\centering
	\begin{minipage}{0.7\linewidth}
\begin{algorithm}[H]
	\caption{Pseudo-code of CEGAN}
	\label{alg:learning_algo}
	\begin{algorithmic}
		\footnotesize
		\STATE {\bfseries Input:} Observational dataset $\mathcal{D}$
		\STATE {\bfseries Output:} CEGAN parameters $(\theta_E, \theta_I, \theta_P, \theta_{D}, \theta_{R})$ \\
		\STATE Initialize $(\theta_E, \theta_I, \theta_P, \theta_{D}, \theta_{R})$ \\
		\REPEAT
		\STATE 1) \textit{Reconstruction network} optimization \\
		\STATE Sample minibatch of $k_{r}$ data and noise samples
		\STATE Update $f_{E}, f_{R}$ using stochastic gradient descent (SGD) with gradient:
		\begin{equation} \nonumber
			\nabla_{\! (\theta_E, \theta_R)}  \frac{1}{k_{r}}  \sum_{i=1}^{k_{r}} \loss_{R}(\wv_{i}, \bar{\wv}_{i})
		\end{equation}
		
		\STATE 2) \textit{Prediction network} optimization \\
		\STATE Sample minibatch of $k_{d}$ data and noise samples
		\STATE Update $D$ using SGD with gradient:
		\begin{equation} \nonumber
			- \nabla_{\! \theta_{D}}  \frac{1}{k_{d}}  \sum_{i=1}^{k_{d}} V(\wv_{i}, \hat{\zv}_{i}, \hat{\yv}_{i})
		\end{equation}
		\STATE Sample minibatch of $k_{g}$ data and noise samples
		\STATE Update $f_{E}, f_I, f_{P}$ using SGD with gradient:
		\begin{equation} \nonumber
			\nabla_{\! (\theta_E, \theta_I, \theta_P)}  \frac{1}{k_{g}}  \sum_{i=1}^{k_{g}} \Big[ V(\wv_{i}, \hat{\zv}_{i}, \hat{\yv}_{i}) + \alpha \loss_{P}(\yv_{i}, \hat{\yv}_{i})\Big]
		\end{equation}
		\UNTIL convergence
	\end{algorithmic}
\end{algorithm}
\end{minipage}
\par \vspace{2mm}
}

We train the reconstruction network ($f_{E}, f_{R}$) by optimizing the following objective in a supervised fashion:
\begin{equation} \nonumber
\underset{(\theta_E,\theta_R)}{\text{minimize}} ~~ \sum_{i=1}^{m} \loss_{R}(\wv_{i}, \bar{\wv}_{i})
\end{equation}
where $\wv_{i} = [\xv_{i}, t_{i}, \yv_{i}]$ and $\bar{\wv}_{i} = [\bar{\xv}_{i}, \bar{t}_{i}, \bar{\yv}_{i}]$.
For the prediction network, we define an empirical value function for the min-max optimization problem in \eqref{eq:minmax_prob}:
\begin{equation} \nonumber
V\!(\wv_{i}, \hat{\zv}_{i},  \hat{\yv}_{i}) \! = \! \log \! D\!\left(\hat{\zv}_{i}, \xv_{i}, t_{i}, \yv_{i} \right)  
+ \log\!\left(1  \!-\!  D\!\left(\zv_{i}, \xv_{i}, t_{i}, \hat{\yv}_{i}\right) \right).
\end{equation}
In addition, we define the following reconstruction loss at the prediction decoder $f_P$: 
\begin{equation} \nonumber
\mathcal{L}_{P}(\yv_{i}, \hat{\yv}_{i}) = \ell(\yv_{i}, \hat{\yv}_{i}),
\end{equation}
where $\ell(\cdot)$ is defined as in \eqref{eq:loss_reconstruction}.
Overall, the discriminator and generator iteratively optimize the following objectives, where $\alpha$ is a trade-off parameter:
\begin{equation}\label{eq:disc_gen_opt}
\begin{split}
&\underset{\theta_{D}}{\text{minimize}}  ~~ - \! \sum_{i=1}^{m} V(\wv_{i}, \hat{\zv}_{i}, \hat{\yv}_{i}) \\
&\underset{(\theta_E, \theta_I, \theta_P)}{\text{minimize}}  ~~ \sum_{i=1}^{m} \Big[ V(\wv_{i}, \hat{\zv}_{i}, \hat{\yv}_{i}) + \alpha \loss_{P}(\yv_{i}, \hat{\yv}_{i})\Big]
\end{split}
\end{equation}

When optimizing CEGAN, the prediction network's loss $\loss_{P}$ is used as a regularizer to improve training compared to using only the GAN loss~\eqref{eq:disc_gen_opt}, and the reconstruction network's loss $\loss_{R}$ drives the learning process. Specifically, we train $(f_E,f_R)$ to minimize $\loss_{R}$ iteratively with the GAN loss~\eqref{eq:disc_gen_opt}, which forces $f_{E}$ to learn a latent mapping $\zv$ that is informative enough to reconstruct $(\xv,\yv,t)$ and, thus, drives $f_{I}$ to favor a meaningful latent structure over a trivial one.

\section{Additional Experiments}
\subsection{Benchmarks}
We compare CEGAN with several cutting-edge methods including logistic regression using treatment as a feature (\textbf{LR-1}), logistic regression separately trained for each treatment assignment (\textbf{LR-2}), $k$-nearest neighbor (\textbf{kNN}) \cite{Crump:08}, Bayesian additive regression trees (\textbf{BART}) \cite{Chipman:10}, causal forests (\textbf{CForest}) \cite{Wager:17}, counterfactual regression with Wasserstein distance (\textbf{CFR$_{\text{WASS}}$}) \cite{Shalit:16}\footnote{\url{https://github.com/clinicalml/cfrnet}}, multi-task Gaussian process (\textbf{CMGP}) \cite{Ahmed:NIPS17b} and Causal Effect VAE (\textbf{CEVAE}) \cite{Louizos:17}\footnote{\url{https://github.com/AMLab-Amsterdam/CEVAE}}. For continuous outcomes, logistic regressions are replaced with least squares linear regressions. We also compare against CEGAN trained only with $\loss_{P}$ (\textbf{CEGAN($\loss_{P}$)}), which is equivalent to a feed-forward network consisting of $f_I$ and $f_{P}$. Consequently, CEGAN($\loss_{P}$) does not exploit adversarial learning and does not account for latent confounders (i.e., the $\zv$ inferred by $f_I$ is not trained to have a meaningful relationship with the data samples $(\xv,t,\yv)$).

\subsection{Simulation Settings}
Unless otherwise specified, we set $\alpha = 1$ in \eqref{eq:disc_gen_opt} and assume a 20-dimensional latent space for $\zv$. A fully-connected network is used for each component of the prediction network (i.e., $f_{E}$, $f_{P}$, $f_I$, and $D$) and a multi-output network is used for the reconstruction network $f_{R}$. Each of these networks comprise $3$ layers, $200$ hidden units in each layer, and ReLU activation functions. The networks are trained using an Adam optimizer with a minibatch size of $64$ and a learning rate of $10^{-4}$. A dropout probability of $0.6$ is assumed, and Xavier and zero initializations are applied for weight matrices and bias vectors, respectively. CEGAN is implemented using \texttt{Tensorflow}.

\subsection{Semi-Synthetic Dataset: TWINS proposed in \cite{Louizos:17}}
\begin{table*}[t!]
	\caption{Comparison of $\sqrt{\epsilon_{\text{PEHE}}}$ and $\epsilon_{\text{ATE}}$ (mean $\pm$ std) on the TWINS dataset with Scenario 1.} \label{Table:TWINS_ALL}
	\begin{center}
		\scriptsize
		\setlength{\tabcolsep}{2.5pt}
		\begin{tabular}{|c|c|c|c|c|c|c|c|c|}
			\hline
			\multirow{3}{*}{\textbf{Method}}&\multicolumn{4}{c|}{$\sqrt{\epsilon_{\text{PEHE}}}$} &\multicolumn{4}{c|}{$\epsilon_{\text{ATE}}$} \\ \cline{2-9} 
			&\multicolumn{2}{c|}{$p=0.1$} &\multicolumn{2}{c|}{$p=0.5$}& \multicolumn{2}{c|}{$p=0.1$} &\multicolumn{2}{c|}{$p=0.5$} \\\cline{2-9} 
			&In-sample	   &Out-sample   &In-sample    &Out-sample    &In-sample	   &Out-sample   &In-sample    &Out-sample   \\ \hline
			LR-1				&0.373$\pm$0.00&0.365$\pm$0.00&0.379$\pm$0.00&0.370$\pm$0.00&0.025$\pm$0.01&0.021$\pm$0.01&0.069$\pm$0.02&0.064$\pm$0.02\\
			LR-2				&0.376$\pm$0.00&0.374$\pm$0.01&0.384$\pm$0.01&0.381$\pm$0.01&0.021$\pm$0.01&0.016$\pm$0.01&0.069$\pm$0.02&0.063$\pm$0.02\\
			kNN   		     	&0.385$\pm$0.01&0.398$\pm$0.01&0.409$\pm$0.01&0.422$\pm$0.01&0.020$\pm$0.02&0.019$\pm$0.02&0.116$\pm$0.03&0.111$\pm$0.03\\
			CForest 	  	    &0.400$\pm$0.30&0.410$\pm$0.26&0.409$\pm$0.33&0.416$\pm$0.27&0.016$\pm$0.01&0.024$\pm$0.01&0.055$\pm$0.02&0.066$\pm$0.02\\
			BART   	     		&0.400$\pm$0.02&0.397$\pm$0.02&0.455$\pm$0.03&0.456$\pm$0.03&0.074$\pm$0.05&0.078$\pm$0.05&0.234$\pm$0.06&0.241$\pm$0.07\\
			CMGP				&0.377$\pm$0.01&0.370$\pm$0.02&0.380$\pm$0.01&0.371$\pm$0.01&0.054$\pm$0.02&0.049$\pm$0.02&0.078$\pm$0.03&0.072$\pm$0.03\\
			CFR$_{\text{WASS}}$ &0.373$\pm$0.00&0.366$\pm$0.00&0.379$\pm$0.01&0.373$\pm$0.01&0.024$\pm$0.02&0.021$\pm$0.02&0.068$\pm$0.03&0.063$\pm$0.03\\
			CEVAE   	   		&\textbf{0.369$\pm$0.00}&0.367$\pm$0.00&0.373$\pm$0.00&0.368$\pm$0.01&\textbf{0.015$\pm$0.01}&0.020$\pm$0.01&0.032$\pm$0.02&0.027$\pm$0.02\\ \hline
			CEGAN				&0.371$\pm$0.00&\textbf{0.364$\pm$0.00}&\textbf{0.372$\pm$0.00}&\textbf{0.366$\pm$0.00}&0.021$\pm$0.01&\textbf{0.016$\pm$0.01}&\textbf{0.030$\pm$0.01}&\textbf{0.026$\pm$0.01}\\ \hline
		\end{tabular}
		\vspace{-3mm}
	\end{center}
\end{table*}

In this subsection, we compare CEGAN against the aforementioned benchmarks using a semi-synthetic dataset that was first proposed in \cite{Louizos:17}. The dataset is based on records of twin births in the USA from 1989-1991 \cite{Almond:05}. Using this real-world dataset, we artificially create a binary treatment such that $t=1$ ($t=0$) denotes being born the heavier (lighter) twin. The binary outcome corresponds to the mortality of each of the twins in their first year. Since we have records for both twins, we treat their outcomes as two potential outcomes, i.e., $\yv(1)$ and $\yv(0)$, with respect to the treatment assignment of being born heavier. To make a semi-synthetic dataset, we choose same-sex twins, discard features that are only available after birth, and focus on cases where both twins have birth weights below 2 kg. Overall, we have a dataset of 10,286 twins with 49 features related to the parents, the pregnancy, and the birth.\footnote{We made every effort to faithfully reproduce the dataset from its source \cite{Almond:05} using the same criteria as in \cite{Louizos:17}, but did not end up with the same number of twins or features.} The mortality rate of the lighter twin ($t=0$) is $21.64\%$ and the heavier twin ($t=1$) is $15.32\%$, which yields an average treatment effect of $-6.32\%$.

For the TWINS dataset whose data generation process is equivalent to what was proposed in \cite{Louizos:17}, we base our treatment assignment on the feature `\texttt{GESTAT10}', which is a categorical value from 0 to 9 representing the number of gestation weeks. (In this experiment, `\texttt{GESTAT}' is discarded; see the description in the manuscript.) We then follow the treatment and noisy proxy generation procedures reported in \cite{Louizos:17}. Specifically, we let $t_{i}|\xv_{i}, z_{i} \sim \Bern(\sigma(w_{o}^{T}\xv + w_{h}(\frac{z}{10}-0.1)))$, where $\sigma(\cdot)$ denotes the sigmoid function, $w_{o}\sim\Norm(0, 0.1\cdot I)$, and $w_{h}\sim\Norm(9, 0.1)$\footnote{Since we have four more features, we did not obtain comparable $\epsilon_{\text{ATE}}$ using $w_{h}\sim \Norm(5, 0.1)$ as reported in~\cite{Louizos:17}. So, we calibrated the mean of $w_{h}$ from 5 to 9 to achieve similar results.}, and we artificially generate noisy proxies by using three randomly flipped replicas of one-hot encoded `\texttt{GESTAT10}' with flipping probability $p$. 
It is worth to highlight that, compared to the TWINS dataset proposed in the manuscript, artificial proxy variables are created based on the gestational age feature and included as additional observed features, and the treatment depends not only on the latent variable but also on these artificial proxies.

The $\sqrt{\epsilon_{\text{PEHE}}}$ and $\epsilon_{\text{ATE}}$ results are reported in Table \ref{Table:TWINS_ALL} for both in-sample and out-of-sample tests. When the proxy noise is relatively small ($p=0.1$), CEGAN and CEVAE achieve comparable performance to other benchmarks. This aligns with the well-known result that three independent views of a latent feature guarantee that it can be recovered~\cite{allman2009identifiability}, so even techniques that do not account for latent confounders can make accurate predictions.
In contrast, when the artificial proxy variables become too noisy to be useful ($p=0.5$), CEGAN and CEVAE achieve comparable performance to each other and outperform the other benchmarks due to their robustness to latent confounders. 
However, the artificially generated treatments $t_{i}$ in this data set are conditioned on both $\xv_{i}$ and $z_{i}$, which is inconsistent with the causal diagram in Figure \ref{sec:causal_effect}(b), where $t_{i}$ only depends on the latent features $z_{i}$. 

\begin{figure*}[t!]
	\centering 
	\includegraphics[width=0.9\linewidth, trim= 0.1 0.1 0.1 0.1]{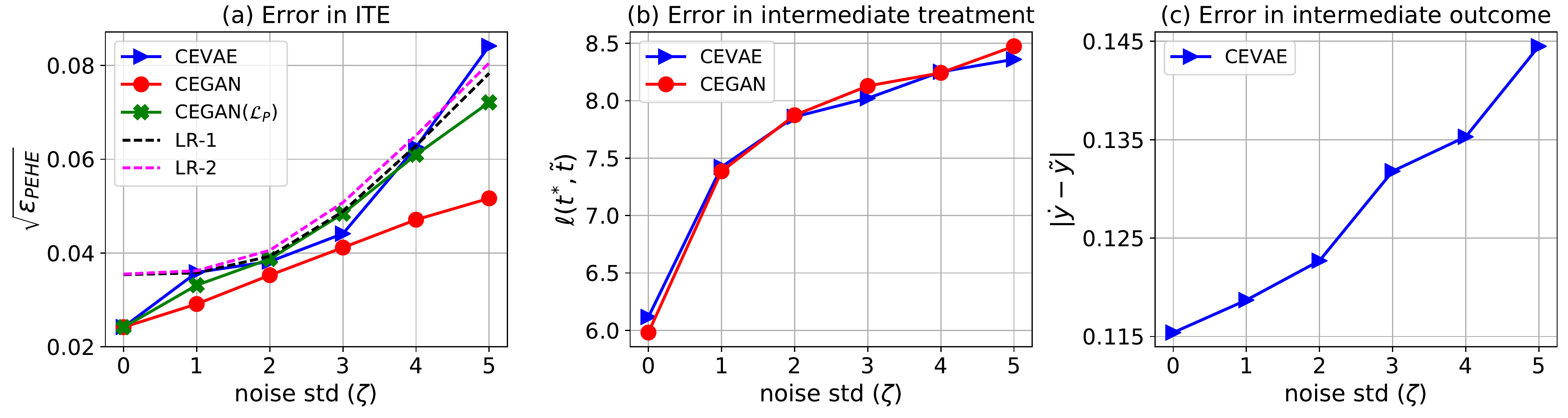}
	\caption{Performance evaluation on the synthetic dataset. The x-axes denote the standard deviation of the noise ($\zeta$) in the proxy mechanism mapping $\zv$ to $\xv$. (a) PEHE vs. $\zeta$ for CEGAN and CEVAE. LR-1, LR-2, and CEGAN($\mathcal{L}_P$) are included for reference. (b) Cross-entropy between $t^{*}$ and $\tilde{t}$. (c) Absolute error between $\dot{\yv}$ and $\tilde{\yv}$.} \label{fig:error_CEVAE} 
\end{figure*} 

\subsection{Synthetic dataset: toy example} \label{sec:experiment_synthetic}
To further illustrate the robustness of CEGAN to latent confounders, we generate a synthetic dataset as follows:
\begin{equation}
\begin{split}
z_{ij} &\sim \Norm(3(\mu-1), 1^2) ~~ \text{for} ~ j=1, \ldots, d_{z} \\
\xv_{i}|\zv_{i} &= \zv_{i} + \nv\\
t_{i}|\zv_{i} &\sim \Bern(\sigma(0.25 \cdot z_{id_{z}})) \\
y_{i}|\zv_{i}, t_{i}  &=  \sigma(\one^{T} \zv_{i} + (2t_{i}-1)),
\end{split}
\end{equation}
where $\mu \sim \Bern(0.5)$, $\nv \sim \Norm(0, \zeta^2 \Iv)$ ($0 \leq \zeta \leq 5$), and $\sigma(\cdot)$ is the sigmoid function. We assume a binary treatment $t \in \{0, 1\}$, a one-dimensional $y \in [0,1]$, and 5-dimensional $\zv$ and $\xv$, i.e., $d_{z}=d_{x}=5$. 

Since we only have access to observations $(\xv, y, t)$, where $\xv$ is a noisy proxy of $\zv$, the above generation process introduces latent confounding between $t$ and $y$ through $\zv$ as illustrated in Figure \ref{fig:causal_model}(b). Without measuring $\zv$, we expect causal inference methods will suffer from confounding bias. In our experiments, we evaluate over a sample size $N=5000$ and average over 50 realizations of the outcomes with the same 64/16/20 train/validation/test splits.

In Figure \ref{fig:error_CEVAE}(a), using out-of-sample tests, we illustrate how $\sqrt{\epsilon_{\text{PEHE}}}$ varies with the standard deviation of the noise ($\zeta$) in the proxy mechanism that maps $\zv$ to $\xv$. The PEHE increases with the noise under all evaluated benchmarks because the conditional entropy of $\zv$ given the proxy $\xv$, i.e., $H(\zv|\xv) = - \E_{p(\xv, \zv)}\left[\log p(\zv|\xv)\right]$, is proportional to $\log \zeta$. However, CEGAN is more robust to the noise than the other benchmarks -- including CEVAE, which considers latent confounders. 

Following the previous discussion regarding its relationship to CEVAE, we believe
that CEGAN performs better than CEVAE because it does not require as many intermediate steps to infer $\zv$ when estimating the ITE \eqref{eq:ITE}. In particular, both CEVAE and CEGAN need to predict an intermediate treatment assignment, i.e., $\tilde{t} \sim q(t|\xv=\xv^{*})$, while CEVAE also needs to predict an intermediate outcome, i.e., $\tilde{y} \sim q(y|\xv=\xv^{*},t=\tilde{t})$, where $(\xv^{*}, t^{*}, y^{*})$ denote the true observations and $(\tilde{t},\tilde{y})$ denote the intermediate predictions. Since the treatment is binary, we adopt the cross-entropy $\ell(t^{*}, \tilde{t})$ defined in \eqref{eq:loss_reconstruction} between $t^{*}$ and $\tilde{t}$ to quantify the error in the predicted intermediate treatment $\tilde{t}$. This error affects both CEVAE and CEGAN. To evaluate the error in predicting the intermediate outcome $\tilde{y}$ we compute the absolute difference between $\dot{y}$ and $\tilde{y}$, i.e., $|\dot{y} - \tilde{y}|$, where $\dot{y} \sim q(y|\xv=\xv^{*}, t=t^{*})$ is the intermediate outcome conditioned on $t^{*}$ instead of $\tilde{t}$. This error only affects CEVAE. In Figure \ref{fig:error_CEVAE}(b) and \ref{fig:error_CEVAE}(c), we show how $\ell(t^{*},\tilde{t})$ and $|\dot{y}-\tilde{y}|$ vary with respect to the standard deviation of the noise ($\zeta$), respectively.

Figure \ref{fig:error_CEVAE}(b) and \ref{fig:error_CEVAE}(c) demonstrate that the intermediate predictions made in both CEGAN and CEVAE become less accurate as the noise increases. We conjecture that this error is accumulated and propagated to the inference of $\zv$ and eventually decreases the accuracy of the ITE estimates. Consequently, since CEVAE requires more intermediate steps to infer $\zv$, it performs worse than CEGAN. 
Note that, we omit error measurements in the latent space because (i) differences between latent variables do not necessarily correspond to the accuracy of predictions based on them and (ii) we cannot directly compare errors in the different latent spaces generated by CEVAE and CEGAN. 
\pagebreak
\end{document}